\documentclass{article}
\pdfoutput=1
\usepackage{arxiv}
\usepackage[utf8]{inputenc}
\usepackage[T1]{fontenc}
\usepackage[colorlinks=true,linkcolor=black,citecolor=CaribbeanCurrent,urlcolor=CaribbeanCurrent]{hyperref}
\usepackage{url}
\usepackage{booktabs}
\usepackage{amsfonts}
\usepackage{amsmath,amssymb}
\usepackage{nicefrac}
\usepackage{microtype}
\usepackage{graphicx}
\usepackage{natbib}
\usepackage{doi}
\usepackage{xcolor}
\usepackage{multirow}
\usepackage{array}
\usepackage{caption}
\usepackage{subcaption}
\usepackage{siunitx}
\usepackage{enumitem}
\usepackage{pifont}
\usepackage[ruled,vlined]{algorithm2e}
\usepackage[most]{tcolorbox}

\definecolor{CaribbeanCurrent}{HTML}{187077}
\definecolor{EarthYellow}{HTML}{E0A458}
\definecolor{PastelCyan}{HTML}{00B4D8}
\definecolor{PastelOrange}{HTML}{F66A0A}

\colorlet{PrimaryColor}{CaribbeanCurrent}
\colorlet{SecondaryColor}{EarthYellow}
\colorlet{LinkColor}{PrimaryColor}
\colorlet{llm}{PastelCyan!30}
\colorlet{REPLCOLOR}{PastelOrange!30}

\newcommand{\RVLM}{\textsc{Rvlm}}
\newcommand{\RRouter}{\textsc{RecursionRouter}}

\title{RVLM: Recursive Vision-Language Models with
	Adaptive Depth}
\date{}

\author{
	\href{https://orcid.org/0009-0006-2489-2922}{\includegraphics[scale=0.06]{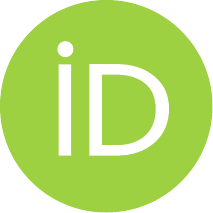}\hspace{1mm}Nicanor Mayumu} \\
	Department of Computer Science\\
	University of Wollongong in Dubai\\
	Dubai Knowledge Park, Dubai, UAE \\
	\texttt{nngwanga@uow.edu.au} \\
	\And
	\href{https://orcid.org/0000-0002-3386-4222}{\includegraphics[scale=0.06]{orcid.pdf}\hspace{1mm}Zeenath Khan} \\
	Department of Computer Science\\
	University of Wollongong in Dubai\\
	Dubai Knowledge Park, Dubai, UAE \\
	\texttt{ZeenathKhan@uowdubai.ac.ae} \\
	\And
	\href{https://orcid.org/0000-0001-7933-2750}{\includegraphics[scale=0.06]{orcid.pdf}\hspace{1mm}Melodena Stephens} \\
	Department of Academic Affairs\\
	Mohammed Bin Rashid School of Government\\
	City Walk, Building B02, Dubai, UAE \\
	\texttt{melodena.stephensb@mbrsg.ac.ae} \\
	\And
	\href{https://orcid.org/0000-0001-6497-1373}{\includegraphics[scale=0.06]{orcid.pdf}\hspace{1mm}Patrick Mukala} \\
	Department of Computer Science\\
	University of Wollongong in Dubai\\
	Dubai Knowledge Park, Dubai, UAE \\
	\texttt{PatrickMukala@uowdubai.ac.ae} \\
	\And
	\href{https://orcid.org/0000-0001-7942-5965}{\includegraphics[scale=0.06]{orcid.pdf}\hspace{1mm} Farhad Oroumchian} \\
	Department of Computer Science\\
	University of Wollongong in Dubai\\
	Dubai Knowledge Park, Dubai, UAE \\
	\texttt{FarhadOroumchian@uowdubai.ac.ae} \\
}

\renewcommand{\shorttitle}{RVLM: Recursive Vision-Language Models}

\hypersetup{
	pdftitle={RVLM: Auditable Recursive Vision-Language Models with Adaptive Depth},
	pdfsubject={cs.CV, cs.AI, cs.CL},
	pdfauthor={Nicanor Mayumu, Zeenath Khan, Melodena Stephens, Patrick Mukala, Farhad Oroumchian},
	pdfkeywords={Vision Language Models, Recursive Reasoning, Adaptive Computation,
		Medical Imaging, REPL, Interpretability, Brain MRI, Chest X-Ray},
}

\begin{document}
	\maketitle
	
	\begin{abstract}
		Medical AI systems face two fundamental limitations. First, conventional vision-language models (VLMs) perform single-pass inference, yielding black-box predictions that cannot be audited or explained in clinical terms. Second, iterative reasoning systems that expose intermediate steps rely on fixed iteration budgets wasting compute on simple cases while providing insufficient depth for complex ones. We address both limitations with a unified framework. RVLM replaces single-pass inference with an iterative generate-execute loop: at each step, the model writes Python code, invokes vision sub-agents, manipulates images, and accumulates evidence. Every diagnostic claim is grounded in executable code, satisfying auditability requirements of clinical AI governance frameworks. RRouter makes iteration depth adaptive: a lightweight controller predicts the optimal budget from task-complexity features, then monitors progress and terminates early when reasoning stalls. We evaluate on BraTS 2023 Meningioma (brain MRI) and MIMIC-CXR (chest X-ray) using Gemini 2.5 Flash without fine-tuning. Across repeated runs, RVLM shows high consistency on salient findings (e.g., mass presence and enhancement) and can detect cross-modal discrepancies between Fluid-Attenuated Inversion Recovery (FLAIR) signal characteristics and segmentation boundaries. On MIMIC-CXR, it generates structured reports and correctly recognises view-specific artefacts. Code: \url{https://github.com/nican2018/rvlm}.
	\end{abstract}
	
	
	\section{Introduction}
	\label{sec:introduction}
	
	The integration of artificial intelligence into clinical radiology has progressed rapidly, with vision-language models (VLMs) demonstrating impressive capabilities in tasks ranging from lesion detection to automated report generation~\citep{singhal2023medpalm, thirunavukarasu2023large}. These models can process multimodal inputs medical images alongside textual prompts and produce diagnostic outputs that increasingly approach clinical expert performance. Yet despite these technical advances, widespread adoption in high-stakes clinical settings remains elusive. Two fundamental limitations persist: the opacity of model reasoning and the inflexibility of computational resource allocation.
	
	\subsection{The Interpretability Challenge}
	
	Current medical VLMs operate predominantly as single-pass inference systems. Given an image and a prompt, they produce a diagnosis or report in one forward pass, revealing none of the intermediate reasoning that led to the conclusion. This black-box nature conflicts directly with the evidentiary standards of clinical medicine~\citep{ghassemi2021clinical}. When a radiologist interprets a study, they can articulate which anatomical structures were examined, what features supported or contradicted each differential diagnosis, and how quantitative measurements informed their assessment. A clinician using an AI system requires the same transparency: the ability to audit, verify, and if necessary, challenge the model's reasoning~\citep{raji2020closing, amini2024foundation}.
	
	The consequences of opaque AI in medicine extend beyond clinician distrust. Regulatory frameworks increasingly mandate transparency. The EU AI Act classifies medical AI as high-risk, requiring systems to provide meaningful explanations of their decisions and enable human oversight. ISO/IEC 42001 and the NIST AI Risk Management Framework similarly emphasise traceability, reproducibility, and auditability. Current VLMs, designed for benchmark performance rather than regulatory compliance, cannot meet these requirements.
	
	\subsection{The Fixed-Budget Problem}
	
	A second limitation compounds the first. Recent work has explored iterative reasoning systems that do expose intermediate steps: chain-of-thought prompting, REPL-based execution, and agentic frameworks that decompose problems into subtasks. These systems improve interpretability by revealing the model's step-by-step reasoning. However, they universally rely on a \emph{fixed iteration budget}, a user-specified \texttt{max\_iterations} hyperparameter applied uniformly across all cases regardless of diagnostic complexity.
	
	This one-size-fits-all approach creates a dilemma. A simple case, a single well-defined meningioma with clear boundaries, may require only three reasoning steps: identify the lesion, measure its dimensions, characterise its appearance. Allocating a fixed budget of twelve iterations wastes computational resources, increases API costs, and adds unnecessary latency. Conversely, a complex case, a multifocal tumour with ambiguous margins, peritumoural oedema, and heterogeneous enhancement \emph{benefits} from additional iterations to resolve uncertainty, consult multiple views, and cross-validate findings. A fixed budget either under-serves complex cases or over-serves simple ones. The optimal number of reasoning steps is a property of the task, not the system, yet existing architectures cannot adapt.
	
	\subsection{Our Contributions}
	
	We address both limitations simultaneously through a unified framework combining two complementary components. \textbf{\RVLM{}} (Recursive Vision-Language Model) provides the interpretability layer, replacing single-pass inference with an iterative generate-execute loop where reasoning unfolds step by step in executable code. \textbf{\RRouter{}} makes iteration depth itself adaptive, predicting case complexity and terminating reasoning when progress stalls.
	
	Specifically, this paper makes the following contributions:
	
	\begin{itemize}[leftmargin=*,itemsep=4pt]
		\item We extend the RLM REPL framework~\citep{zhang2025rlm} with first-class image support, introducing primitives for visual reasoning: \texttt{context\_images} for maintaining multimodal working memory, \texttt{describe\_image} for generating natural language descriptions of regions of interest, \texttt{llm\_query\_with\_images} for visual question answering, and programmatic image manipulation functions (crop, enhance, difference maps). The model accumulates evidence across iterations, building a verifiable reasoning trace where every claim corresponds to executable code.
		
		\item We introduce a pre-flight complexity estimator that predicts optimal iteration depth from task features (label entropy, tumour volume, sub-region count, presence of tiny regions) and a per-iteration stall detector that monitors reasoning progress. This treats recursion depth as a latent variable inferred from the input, eliminating the fixed-budget hyperparameter entirely.
		
		\item We provide an explicit mapping from \RVLM{}'s mechanisms to the requirements of clinical AI governance. The system's executable traces directly satisfy EU AI Act transparency obligations, ISO/IEC 42001 traceability requirements, and NIST AI RMF principles of explainability and human oversight. We argue that interpretability must be architected in, not added post-hoc.
		
		\item Recognising that executable code is not a clinician-friendly output, we implement a secondary LLM pass that converts \RVLM{}'s reasoning traces into formal, structured patient reports in PDF format. This abstraction layer maintains the verifiability of the underlying code while presenting information in the familiar style of radiology documentation.
		
		\item We evaluate the combined system on two clinically distinct benchmarks without task-specific fine-tuning. On BraTS 2023 Meningioma (brain MRI with four modalities and segmentation overlays), \RVLM{} achieves 90\% consistency on high-salience findings across 16 independent runs and autonomously detects cross-modal discrepancies absent from single-pass baselines. On MIMIC-CXR (chest radiography with PA/Lateral/AP views), the same infrastructure generates structured Findings and Impression reports, correctly recognising view-specific artefacts and producing clinically coherent documentation.
	\end{itemize}
	
	\subsection{Precision-Latency-Complexity Tradeoff}
	
	\RVLM{} explicitly prioritises \emph{verifiability} and \emph{transparency} over raw inference speed. The system is designed not to replace real-time screening tools, but to serve as a complementary technology for high-stakes scenarios where reasoning matters as much as results: second-opinion systems for complex cases, research validation pipelines, pre-operative planning, and radiology education. \RRouter{} recovers efficiency where possible by matching computational investment to case difficulty, ensuring that the system's latency reflects the complexity of the diagnostic task.
	
	\subsection{Paper Organisation}
	
	The remainder of this paper is organised as follows. Section~\ref{sec:related} reviews related work in medical VLMs, iterative reasoning systems, and adaptive computation. Section~\ref{sec:methodology} details the \RVLM{} architecture, including the REPL environment, vision primitives, and state management. Section~\ref{sec:router} introduces \RRouter{} and its complexity prediction and stall detection mechanisms. Section~\ref{sec:trust} maps the architecture to regulatory trust frameworks, providing a concrete pathway to compliance. Section~\ref{sec:experiments} presents experimental setup and results on BraTS 2023 Meningioma and MIMIC-CXR. Section~\ref{sec:discussion} analyses clinical implications, limitations, and future directions. Section~\ref{sec:conclusion} concludes.
	\section{Related Work}
	\label{sec:related}
	
	\paragraph{Vision-Language Models (VLMs).}
	Large-scale VLMs such as GPT-4o~\citep{openai2024gpt4o}, 
	Gemini~\citep{team2024gemini}, Claude~3.5~\citep{anthropic2024claude35},
	and the LLaVA family~\citep{liu2023llava,liu2024llavaone} achieve impressive
	performance across visual understanding, captioning, and multimodal reasoning~\cite{mayumu2025omniv2x}.
	These models predominantly rely on single-pass inference; the entire visual
	and textual reasoning must be compressed into one forward pass.
	Recent efforts introduce visual chain-of-thought~\citep{zhang2024cocot,wang2024visualcot}
	and multi-turn strategies, yet these approaches lack persistent state
	management and do not support programmatic control over visual processing.
	
	\paragraph{Recursive and iterative language model frameworks.}
	The Recursive Language Model (RLM) paradigm~\citep{zhang2025rlm} embeds a
	language model inside a persistent REPL, enabling recursive text manipulation
	and long-horizon reasoning through stateful code execution.
	Related iterative frameworks include ReAct~\citep{yao2023reactsynergizingreasoningacting},
	Reflexion~\citep{shinn2023reflexion}, THREAD~\citep{schroeder2025threadthinkingdeeperrecursive},
	and LADDER~\citep{simonds2025ladderselfimprovingllmsrecursive}.
	None natively incorporates vision capabilities or supports recursive image
	manipulation.
	
	\paragraph{Code-augmented and programmatic visual reasoning.}
	ViperGPT~\citep{surismenon2023vipergpt} pioneered program synthesis for
	visual QA by composing modular visual primitives.
	CodeAct~\citep{wang2024executablecodeactionselicit}, Visual
	Program~\citep{hu2024visualprogram}, and Visual
	Chain-of-Code~\citep{chen2025vcoc} further demonstrated that executable
	code reduces hallucination in vision tasks.
	\RVLM{} extends this direction with (1)~a \emph{persistent, stateful REPL}
	enabling iterative refinement, and (2)~\emph{recursive sub-VLM calls} on
	arbitrary image subsets.
	Code-as-thought approaches~\citep{chen2022program,gao2023pal} ground
	reasoning in verifiable computational steps; our REPL loop inherits this
	tradition and extends it to multimodal, multi-step medical imaging.
	
	\paragraph{Adaptive computation.}
	Early-exit networks~\citep{teerapittayanon2016branchynet,schwartz2020right}
	and mixture-of-depths~\citep{raposo2024moe} allocate variable compute per
	instance.
	PonderNet~\citep{banino2021pondernet} treats computation steps as a latent
	variable trained end-to-end.
	\RRouter{} applies analogous ideas to LLM iteration depth, using
	domain-specific task features rather than learned internal states, 
	a practical advantage when outcome labels are scarce.
	
	\paragraph{Medical vision-language models and trustworthy AI.}
	Med-PaLM~M~\citep{tu2024medpalm}, LLaVA-Med~\citep{li2024llavamed},
	Med-Gemini~\citep{tu2025medgemini}, and XrayGLM~\citep{chen2024xrayglm}
	demonstrate strong radiology performance but inherit the single-pass
	limitation.
	Trustworthy medical AI research emphasises explainability~\citep{selvaraju2017gradcam}
	and uncertainty quantification~\citep{obermeyer2019dissecting}; few works
	combine these with deep iterative visual reasoning.
	\RVLM{} grounds every inference step in auditable Python code.
	
	\paragraph{Radiology report generation.}
	Automated chest radiograph report generation has evolved from
	template-based~\citep{jing2017automatic} to
	transformer-based~\citep{chen2020generating,miura2021improving} and
	VLM-driven~\citep{hyland2024maira,chen2024chexagent} approaches.
	\RVLM{} differs by not requiring task-specific fine-tuning: the same REPL
	infrastructure, applied to brain MRI and chest X-ray through modality-aware
	prompt templates, adapts without weight updates.
	
	\paragraph{BraTS challenge and brain tumour analysis.}
	The BraTS Challenge~\citep{menze2015brats,labella2023brats,bakas20242024}
	has served as the canonical brain tumour benchmark for over a decade.
	\RVLM{} is, to our knowledge, the first approach to apply recursive,
	code-augmented vision-language reasoning to this challenge, moving beyond
	segmentation toward interpretable iterative diagnostic reasoning.

	\section{RVLM Methodology}
	\label{sec:methodology}
	
	\subsection{Architectural Paradigm}
	
\begin{figure}[h]
		\centering
		\includegraphics[width=0.9\textwidth]{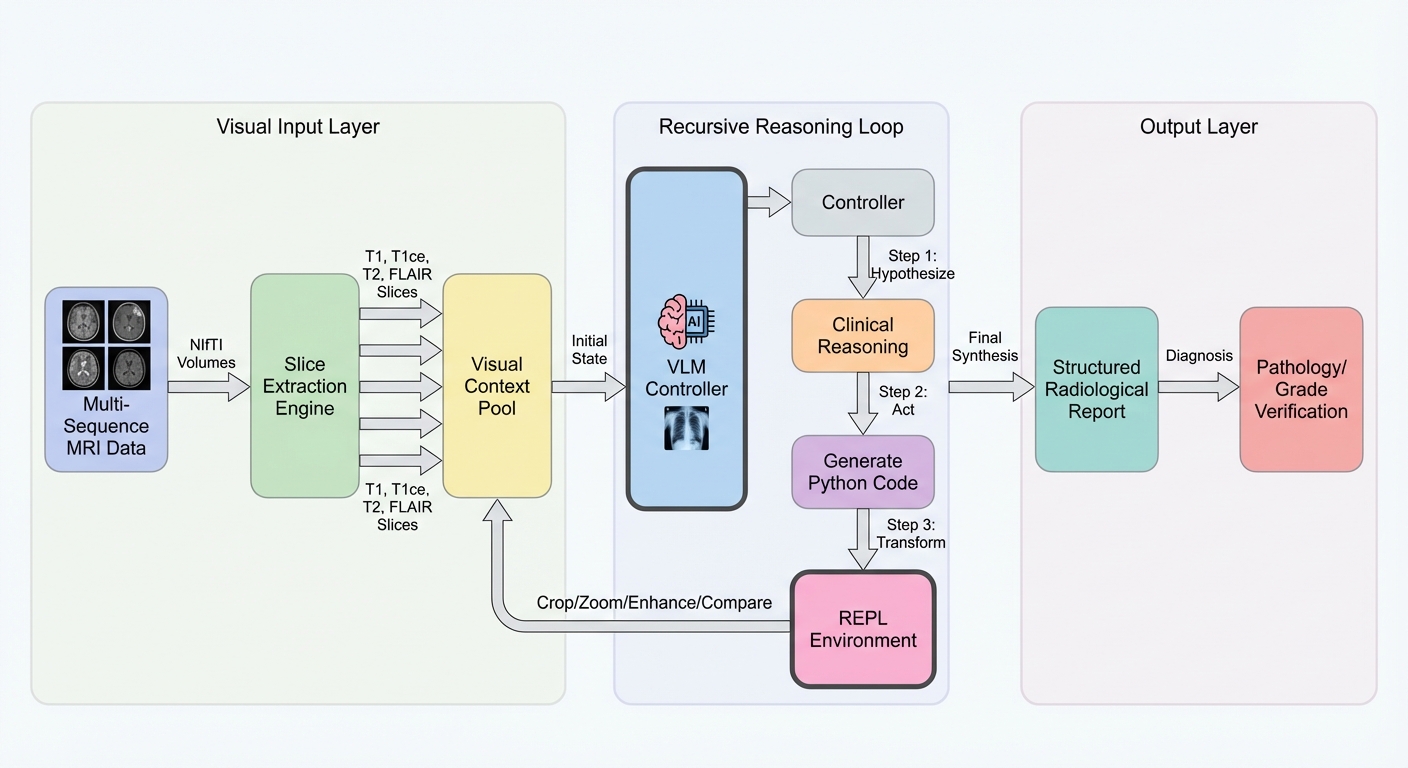}
		\caption{\textbf{\RVLM{} System Architecture.}
			The core VLM $\mathcal{M}_V$ operates within a persistent Python
			REPL environment $\mathcal{E}_V$ that maintains state across
			iterations.
			The environment provides first-class image storage
			(\texttt{context\_images}), vision-specific query primitives, and
			full programmatic image manipulation.
			At each iteration~$t$, the model generates executable code, which
			is executed in the REPL; outputs are appended to the message history.
			The loop terminates upon emission of a \texttt{FINAL} or
			\texttt{FINAL\_VAR} signal.}
		\label{fig:rvlm-overview}
	\end{figure}

	The \RVLM{} framework extends the Recursive Language Model
	paradigm~\citep{zhang2025rlm} to multimodal visual reasoning.
	Our central hypothesis is that image-based diagnostic reasoning, like textual
	analysis, benefits from recursive decomposition: a model capable of
	programmatically inspecting, manipulating, and recursively querying medical
	images across iterative cycles produces more verifiable and clinically
	auditable results than monolithic architectures constrained to single-pass
	inference.
	\RVLM{} achieves this by embedding a vision-capable language model within a
	persistent visual REPL environment, enabling stateful, multi-step reasoning
	with procedural transparency.
	The overall system architecture is illustrated in
	Figure~\ref{fig:rvlm-overview}.
	
	\subsection{System Components}
	
	The \RVLM{} system comprises three tightly integrated components:
	\begin{enumerate}[leftmargin=*,itemsep=2pt]
		\item \textbf{Vision-capable root LM ($\mathcal{M}_V$):} A multimodal
		foundation model (e.g., Gemini~2.5~Flash) capable of processing
		interleaved text and image inputs.
		$\mathcal{M}_V$ serves dual roles: primary controller that generates
		executable code, and backend for recursive sub-VLM calls.
		
		\item \textbf{Visual REPL environment ($\mathcal{E}_V$):} An extended
		Python REPL that maintains persistent state across iterations.
		$\mathcal{E}_V$ treats images as first-class objects and exposes
		vision-specific utility functions, creating a \emph{visual working
			memory} for intermediate analysis artefacts.
		
		\item \textbf{Recursive vision calls:} A mechanism enabling sub-VLM
		invocations on arbitrary subsets of images, facilitating multi-scale
		inspection, cross-modal comparison, and iterative refinement within
		a unified reasoning loop.
	\end{enumerate}
	
	The formal execution procedure is detailed in Algorithm~\ref{alg:rvlm-call}.
	Only the initial user message is multimodal (text + images); subsequent
	messages are text-only, with visual data accessed exclusively through REPL
	primitives, reducing per-iteration API cost by ${\approx}$70-80\%.
	
	\begin{algorithm}[h]
		\caption{\RVLM{}: Recursive Vision-Language Model}
		\label{alg:rvlm-call}
		\KwIn{text prompt $P$, images $\mathbf{I} = \{I_1, \ldots, I_k\}$, max iterations $T$}
		\KwOut{response $Y$}
		$\mathcal{E}_V \gets \texttt{InitREPL}(P)$\;
		$\mathcal{E}_V.\texttt{context\_images} \gets \texttt{Encode}(\mathbf{I})$\;
		$\mathcal{E}_V \gets \texttt{InjectVisionFunctions}(\mathcal{E}_V, \mathbf{I})$\tcp*[r]{\scriptsize \texttt{describe\_image}, \texttt{llm\_query\_with\_images}}
		$\texttt{hist} \gets [\texttt{SystemPrompt}, \texttt{Metadata}(\mathcal{E}_V)]$\;
		\For{$t = 0$ \KwTo $T-1$}{
			\eIf{$t = 0$}{
				$\texttt{hist} \gets \texttt{hist} \,\Vert\, \texttt{MultimodalMessage}(P, \mathbf{I})$\;
			}{
				$\texttt{hist} \gets \texttt{hist} \,\Vert\, \texttt{TextMessage}(t)$\;
			}
			$\texttt{code} \gets \mathcal{M}_V(\texttt{hist})$\;
			$(\mathcal{E}_V, \texttt{stdout}) \gets \texttt{REPL}(\mathcal{E}_V, \texttt{code})$\;
			$\texttt{hist} \gets \texttt{hist} \,\Vert\, \texttt{code} \,\Vert\, \texttt{stdout}$\;
			\If{$\mathcal{E}_V[\texttt{FINAL}]$ \textnormal{is set}}{
				\Return $\mathcal{E}_V[\texttt{FINAL}]$\;
			}
		}
		\Return $\texttt{DefaultAnswer}(\texttt{hist}, \mathcal{M}_V)$\;
	\end{algorithm}
	
	\subsection{Visual REPL Extensions}
	\label{sec:visual-repl}
	
	The \RVLM{} REPL extends the standard RLM environment with three additions
	injected at initialisation via \texttt{\_inject\_image\_support()}.
	
	\subsubsection{Image Context}
	A \texttt{context\_images} list is added to the REPL namespace.
	Each image is represented as a dictionary with keys: \texttt{data}
	(base64-encoded string or URL), \texttt{media\_type} (MIME type), and
	\texttt{detail} (resolution hint).
	This abstraction enables image reference by index, eliminating raw pixel
	data from the message history and significantly reducing token usage.
	
	\subsubsection{Vision Query Functions}
	\begin{itemize}[leftmargin=*,itemsep=2pt]
		\item \texttt{describe\_image(index, prompt)}: Sends a single image
		to a sub-VLM and returns a textual description.
		
		\item \texttt{llm\_query\_with\_images(prompt, image\_indices,
			image\_sources)}: Constructs multimodal messages and dispatches to
		a sub-VLM, enabling selective querying, derived image analysis, and
		cross-modal comparison.
		
		\item \texttt{llm\_query\_batched\_with\_images(prompts, image\_indices,
			image\_sources)}: Batched variant pairing each prompt with a
		corresponding image specification for parallel multi-image analysis.
	\end{itemize}
	
	\subsubsection{Programmatic Image Manipulation}
	The REPL grants full access to Python libraries (PIL, NumPy, Matplotlib)
	for image transformation between iterations: cropping regions of interest,
	applying contrast enhancement (CLAHE), computing difference maps
	(e.g., $\text{T1ce} - \text{T1n}$), and extracting quantitative
	measurements (mean intensities, region areas).
	This composability code execution producing new images for recursive
	vision calls distinguishes \RVLM{} from standard VLM pipelines and
	enables quantitative verification of visual hypotheses.
	
	\subsection{Modality-Aware Decomposition}
	\label{sec:modality-decomposition}
	
	A key design principle is \emph{modality-aware decomposition}: the prompt
	instructs \RVLM{} to issue one targeted \texttt{describe\_image()} call per
	imaging modality (or per X-ray view), with a clinically specific sub-prompt
	tailored to what that modality best reveals.
	
	For \emph{brain MRI}: T1-native is described with emphasis on lesion
	laterality and anatomical boundaries; T1-contrast with focus on enhancement
	pattern and dural tail; T2-weighted with emphasis on oedema extent; T2-FLAIR
	with attention to peritumoral signal.
	For \emph{chest radiography}: the PA view uses a five-item systematic
	checklist (lung fields, cardiomediastinal contour, pleura, bones, diaphragm);
	the Lateral view focuses on retrosternal space and posterior costophrenic
	angles; the AP portable view includes a note on cardiac magnification
	artefact.
	
	This decomposition prevents attention dilution from simultaneous multi-image
	processing, and ensures each modality's contribution is captured as a named,
	auditable REPL variable.
	
	\subsection{Cross-Modal Synthesis}
	\label{sec:cross-modal}
	
	After per-modality descriptions, the model performs a cross-modal assessment
	via \texttt{llm\_query\_with\_images()} with diagnostically paired images
	(e.g., T1ce and FLAIR for brain MRI; PA and Lateral for chest X-ray).
	The individual descriptions and the cross-modal query are assembled into an
	\texttt{evidence} string (each truncated to avoid context overflow) that is
	passed to a synthesis \texttt{llm\_query()} call producing a formal,
	structured radiology report in plain prose.
	
	\subsection{Ground-Truth Grounding}
	\label{sec:gt-grounding}
	
	For BraTS, where voxel-level segmentation masks are available, \RVLM{}
	renders the segmentation overlay as an additional image alongside the four
	modality slices.
	Mask statistics (volume in cc and percentage share for each sub-region) are
	included in the text prompt, enabling the model to compare its visual
	impression against quantitative ground truth and reason explicitly about
	agreements and discrepancies.
	
	\subsection{Clinical PDF Reporting Sub-Agent}
	\label{sec:report}
	
	While \RVLM{}'s REPL log provides full auditability for technical reviewers,
	clinicians require formatted clinical documents.
	We implement a \emph{clinical PDF reporting sub-agent} that runs after the
	REPL loop and converts the \RVLM{} synthesis output into a formal,
	printable patient report.
	
	The sub-agent operates in two stages:
	\begin{enumerate}[leftmargin=*,itemsep=2pt]
		\item \textbf{Structured extraction:} A secondary LLM call parses the
		free-text synthesis into named JSON fields.
		For \emph{neuroradiology}: Location, Sub-region Analysis, Mass
		Effect, Key Imaging Features, and GT Agreement.
		For \emph{chest radiography}: Lungs, Cardiac Silhouette, Pleural
		Spaces, Bones \& Support Devices, and Impression.
		
		\item \textbf{PDF generation:} A \LaTeX{} template is populated with
		the extracted sections, an execution statistics block (wall-clock
		time, token usage, iterations, sub-calls), and a mandatory AI
		disclaimer footer, then compiled to PDF via \texttt{pdflatex}.
	\end{enumerate}
	
\begin{figure}[h]
		\centering
		\fbox{\includegraphics[width=0.75\linewidth]{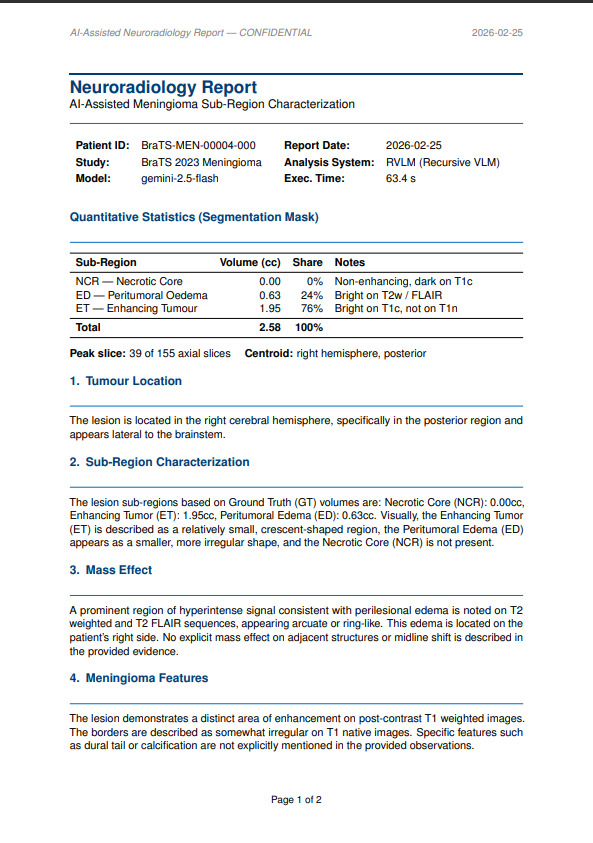}}
		\caption{\textbf{Auto-generated clinical PDF report-BraTS-MEN-00004-000.}
			Five labelled sections, mask-statistics table, and AI disclaimer footer.
			Intended for clinical review, not autonomous diagnosis.}
		\label{fig:brats_report}
	\end{figure}

\begin{figure}[h]
		\centering
		\fbox{\includegraphics[width=0.75\linewidth]{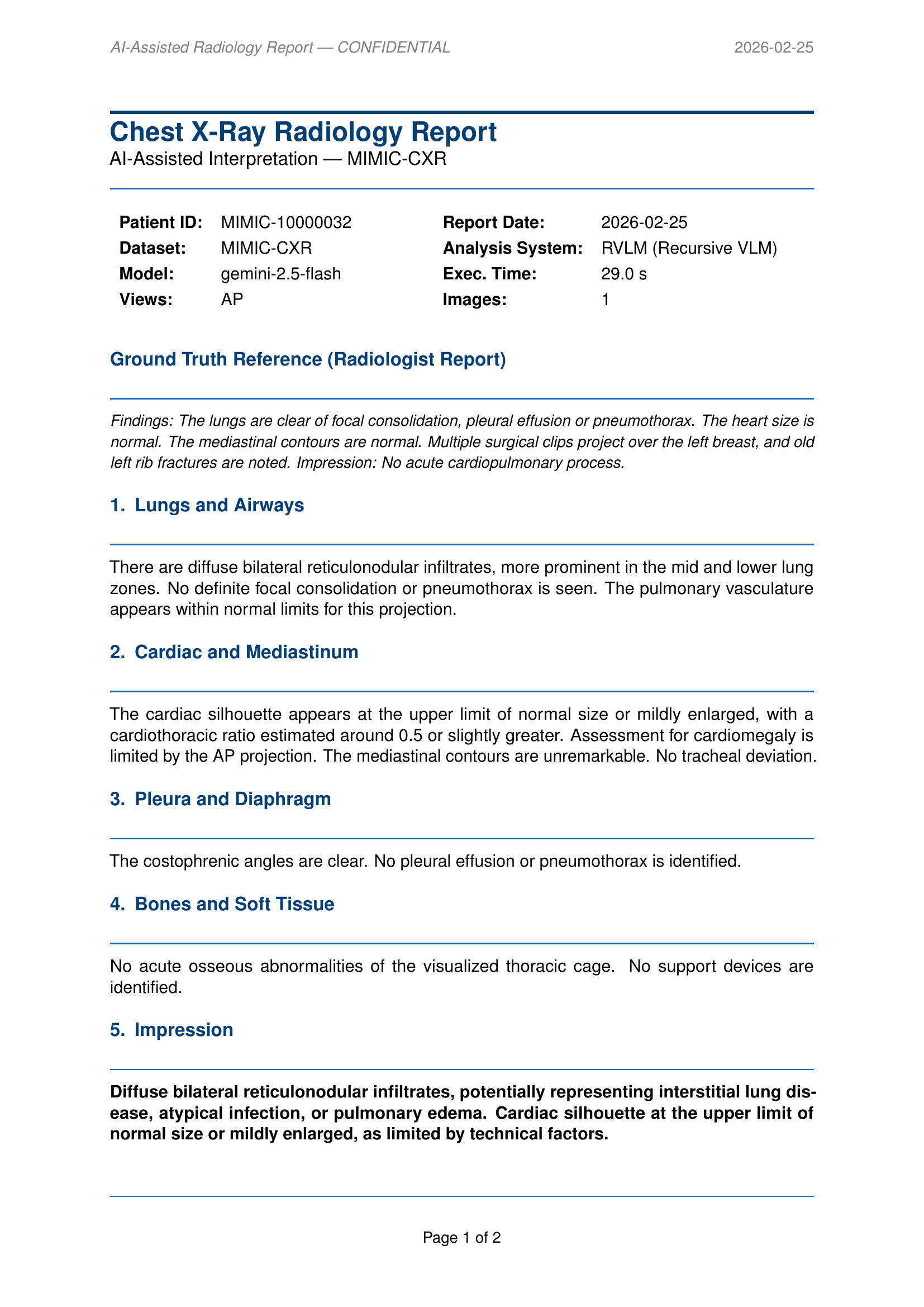}}
		\caption{\textbf{Auto-generated clinical PDF report-MIMIC-CXR subject 10000032.}
			Five clinical sections, a Ground Truth Reference block for radiologist
			comparison, and an AI disclaimer.
			The model correctly identifies the AP portable projection and qualifies
			its cardiac size estimate accordingly.}
		\label{fig:mimic_report}
	\end{figure}

	The CXR variant includes a side-by-side Ground Truth Reference section
	for direct radiologist comparison.
	Figures~\ref{fig:brats_report} and~\ref{fig:mimic_report} show example
	outputs.
	
	\subsection{Interpretability and Verifiability}
	\label{sec:interpretability}
	
	\RVLM{} addresses the black-box nature of conventional VLMs by grounding
	reasoning in executable procedures, providing three complementary levels of
	interpretability.
	\emph{Procedural transparency}: every visual biomarker identification is
	realised through code that isolates, quantifies, or highlights the relevant
	region.
	\emph{Visual provenance}: intermediate visualisations (heatmaps, contours,
	annotated crops) link high-level diagnostic statements to specific pixel
	coordinates.
	\emph{Scientific grounding}: the model tests visual hypotheses by generating
	crops, performing contrast subtraction, or computing derived maps and the
	entire trajectory can be replayed identically given the same inputs.

	\section{\RRouter{}: Adaptive Iteration Depth}
	\label{sec:router}
	
	\subsection{Motivation}
	
	The iteration budget \texttt{max\_iterations} is a fixed hyperparameter in
	standard RLM/RVLM deployments.
	This is inefficient in both directions: wasteful for simple single-region
	cases and potentially insufficient for complex multi-region tumours.
	We propose treating recursion depth as a \emph{latent variable} determined
	by task complexity analogous to how PonderNet~\citep{banino2021pondernet}
	treats computation steps, but using interpretable domain features rather than
	learned internal states.
	
	\subsection{Complexity Features}
	
	Given a segmentation mask, we extract four scalar features:
	\begin{enumerate}[leftmargin=*,itemsep=2pt]
		\item \textbf{Label entropy} $H$ (bits): Shannon entropy of the normalised
		volume distribution across NCR, ED, ET:
		\begin{equation}
			H = -\sum_{k} p_k \log_2 p_k, \quad p_k = \frac{v_k}{\sum_j v_j},
			\label{eq:entropy}
		\end{equation}
		where $v_k$ is the volume (cc) of sub-region $k$.
		$H = 0$ when one region dominates; $H_{\max} = \log_2 3 \approx 1.585$
		when all three are equal.
		
		\item \textbf{Total tumour volume} $V$ (cc): larger tumours require more
		analytical steps to characterise.
		
		\item \textbf{Present sub-region count} $R \in \{1,2,3\}$: number of
		sub-regions with volume~$>$~0.01~cc.
		
		\item \textbf{Tiny region indicator} $T \in \{0,1\}$: whether any
		sub-region has volume $\in (0, 0.5)$~cc hard to characterise visually.
	\end{enumerate}
	
	\subsection{Complexity Score and Budget}
	
	The four features are combined into a composite score $s \in [0,1]$:
	\begin{equation}
		s = 0.35 \cdot \frac{H}{H_{\max}}
		+ 0.30 \cdot \min\!\left(\frac{V}{50},\, 1\right)
		+ 0.25 \cdot \frac{R}{3}
		+ 0.10 \cdot T.
		\label{eq:score}
	\end{equation}
	The weight vector $(0.35, 0.30, 0.25, 0.10)$ encodes domain priors: label
	entropy is the strongest predictor of analytical difficulty, followed by
	tumour volume and structural complexity.
	These weights are currently hand-set; Section~\ref{sec:future-directions}
	describes how they can be learned from outcome data.
	
	The score maps to an iteration budget via a piecewise-constant function:
	\begin{equation}
		n^*(s) = \begin{cases}
			3 & s < 0.25 \\
			4 & 0.25 \le s < 0.45 \\
			5 & 0.45 \le s < 0.65 \\
			6 & s \ge 0.65
		\end{cases}
		\label{eq:budget}
	\end{equation}
	
	Table~\ref{tab:router} shows \RRouter{} scores for representative cases
	spanning the complexity spectrum.
	
	\begin{table}[h]
		\centering
		\caption{\RRouter{} complexity scoring for representative BraTS-MEN
			cases. $H$ = label entropy (bits), $V$ = total volume (cc),
			$R$ = sub-region count, $T$ = tiny region flag,
			$s$ = composite score, $n^*$ = recommended budget.}
		\label{tab:router}
		\small
		\begin{tabular}{lcccccc}
			\toprule
			Case description & $H$ & $V$ & $R$ & $T$ & $s$ & $n^*$ \\
			\midrule
			Single ET, small (00008-000) & 0.00 & 9.83  & 1 & 0 & 0.14 & 3 \\
			Single ET, large             & 0.00 & 45.0  & 1 & 0 & 0.35 & 4 \\
			Two regions (ET+ED)          & 1.00 & 20.0  & 2 & 0 & 0.58 & 5 \\
			Three regions, high volume   & 1.50 & 30.0  & 3 & 1 & 0.86 & 6 \\
			Three regions, tiny NCR      & 1.10 & 18.0  & 3 & 1 & 0.73 & 6 \\
			\bottomrule
		\end{tabular}
	\end{table}
	
	\subsection{Per-Iteration Stall Detection}
	
	Beyond the pre-computed budget, \RRouter{} monitors each iteration for
	\emph{productivity}: an iteration is productive if it made at least one
	sub-LM call \emph{or} produced meaningful stdout ($\ge 20$ characters)
	\emph{or} created new non-trivial REPL variables.
	When two consecutive iterations are unproductive and the current iteration
	index exceeds the recommended budget, the router terminates early.
	
	If the model has already computed a \texttt{report} variable in the REPL
	environment but failed to emit \texttt{FINAL\_VAR}, the framework recovers
	the variable directly from \texttt{locals} rather than synthesising a generic
	fallback, preventing a common failure mode where correct content is
	discarded in favour of boilerplate.
	
	\subsection{Integration with the RVLM Loop}
	
	\RRouter{} is an optional, backward-compatible addition.
	When \texttt{router=None}, the loop falls back to fixed
	\texttt{max\_iterations}.
	Algorithm~\ref{alg:router-loop} shows the unified RVLM + \RRouter{} loop.
	
	\begin{algorithm}[h]
		\caption{\RVLM{} Completion Loop with \RRouter{}}
		\label{alg:router-loop}
		\KwIn{prompt $p$, images $\mathcal{I}$, router $\rho$ (optional), ceiling $T$}
		\KwOut{answer $Y$}
		$n \gets \rho.\texttt{recommended\_max}()$ if $\rho \neq \emptyset$ else $T$\;
		Inject $\mathcal{I}$ into REPL; build $H_0$\;
		\For{$i = 0$ \KwTo $n-1$}{
			$r_i \gets \mathcal{M}_V(H_i)$; execute REPL block; update \texttt{locals}\;
			\If{\texttt{FINAL\_VAR}$(v)$ or \texttt{FINAL}$(t)$ in $r_i$}{
				\Return \texttt{locals}[$v$] or $t$\;
			}
			Append $r_i$ + REPL output to $H_{i+1}$\;
			\If{$\rho \neq \emptyset$ \textbf{and} $\neg\,\rho.\texttt{should\_continue}(i, r_i, \texttt{locals})$}{
				\Return \texttt{locals}[``report''] if present, else $\texttt{DefaultAnswer}(H_{i+1})$\;
			}
		}
		\Return \texttt{locals}[``report''] if present, else $\texttt{DefaultAnswer}(H_n)$\;
	\end{algorithm}
	
	\subsection{Cost Efficiency}
	
	For BraTS-MEN-00008-000 (single-region, $s=0.14$), the effective budget is
	3 iterations versus the prior fixed ceiling of~12 a four-fold reduction
	in worst-case cost.
	For complex three-region cases ($s \ge 0.65$), the router \emph{increases}
	the budget by up to three iterations, ensuring sufficient analytical depth.
	The system thus achieves simultaneous improvement in both precision (complex
	cases) and efficiency (simple cases).

	\section{\RVLM{} as a Trust-by-Design Architecture}
	\label{sec:trust}
	
	High-risk AI systems in clinical settings are subject to regulatory
	requirements that extend well beyond predictive performance.
	The EU AI Act (Regulation (EU) 2024/1689)~\citep{euaiact2024} mandates
	transparency (Art.~13), human oversight (Art.~14), robustness (Art.~15),
	and quality management (Art.~17).
	ISO/IEC~42001~\citep{iso42001} requires documented AI governance including
	impact assessment, traceability, and audit.
	The NIST AI RMF~\citep{nist_rmf} operationalises these through
	\textit{Govern}, \textit{Map}, \textit{Measure}, and \textit{Manage}
	functions centred on trustworthiness.
	
	While these frameworks specify what high-risk clinical AI systems must demonstrate, they often leave open the question of how such requirements are operationalised in practice at the level of model behaviour. RVLM addresses this gap by translating governance-level trust requirements into concrete technical mechanisms: enforced stepwise reasoning, executable logging, cross-modal verification, and inspectable intermediate artefacts. In doing so, RVLM complements trust frameworks by producing a procedural, reproducible evidence chain that supports post-hoc audit, clinician oversight, and documentation beyond predictive performance. We do not claim regulatory compliance; we show how design choices can generate audit artefacts that support transparency, oversight, and documentation obligations.
	
	\subsection{Trust Dimension Mapping}
	
	Table~\ref{tab:trust} maps each regulatory trust dimension to the specific
	\RVLM{} mechanism that operationalises it.
	
	\begin{table}[h]
		\centering
		\caption{Trust-by-design mapping: regulatory dimension $\rightarrow$
			\RVLM{} mechanism $\rightarrow$ governance implication.}
		\label{tab:trust}
		\small
		\begin{tabular}{p{2.4cm}p{4.4cm}p{4.6cm}}
			\toprule
			\textbf{Trust Dimension} & \textbf{\RVLM{} Mechanism} &
			\textbf{Governance Implication} \\
			\midrule
			Procedural Transparency
			& Stepwise REPL reasoning; one analysis step per iteration with stored intermediate variables and execution logs.
			& Clinicians audit the reasoning path, and clinician-readable PDF report with explicit AI disclaimer; supports EU AI Act Art.~13 \\
			\addlinespace
			Auditability \& Traceability
			&Complete REPL execution log including generated code, sub-VLM calls, stdout outputs, and stored variables; runs are reproducible given identical inputs and model weights
			& Full post-hoc audit trail; satisfies ISO/IEC~42001
			audit requirements \\
			\addlinespace
			Reliability \& Verification
			& Cross-modal verification: independent per-modality
			descriptions, followed by explicit discrepancy detection across modalities
			& Reduces single-source hallucination; mirrors clinical
			cross-referencing practice \\
			\addlinespace
			Human-in-the-Loop
			& Inspectable intermediate states; clinical PDF with AI
			disclaimer; human-readable code
			& Clinician retains decision authority; supports EU AI
			Act Art.~14 \\
			\addlinespace
			Explainability
			& Evidence chain linking pixel-level observations, intermediate REPL variables, and final diagnostic synthesis
			& Reasoning traceable from pixel observation to clinical
			conclusion \\
			\addlinespace
			Adaptive Efficiency
			& \RRouter{} matches compute to case complexity; stall
			detection prevents wasted iterations
			& Enables cost-effective deployment at scale without
			sacrificing depth on complex cases \\
			\bottomrule
		\end{tabular}
	\end{table}
	
	\subsection{Auditability and Traceability}
	
	\RVLM{}'s REPL log records: the exact prompt at each iteration, the generated
	code, the REPL output (including sub-VLM responses with timing and token
	usage), and the final synthesis inputs.
	Because the REPL is a deterministic Python interpreter, any run can be
	replayed exactly given the same model weights, inputs, and REPL state.
	This provides a stronger auditability guarantee than attention-weight
	visualisation (e.g., Grad-CAM~\citep{selvaraju2017gradcam}), which gives a
	post-hoc approximation without a reproducible execution trace.
	
	\subsection{Human-in-the-Loop Review}
	
	\RVLM{} supports human oversight at three levels:
	\begin{enumerate}[leftmargin=*,itemsep=2pt]
		\item \textbf{Variable level:} A clinician or engineer can inspect any
		named REPL variable (\texttt{t1c\_desc}, \texttt{cross\_q}) to
		understand exactly what the model observed.
		\item \textbf{Report level:} The clinical PDF sub-agent produces a
		structured document labelled ``AI-generated'' with a mandatory
		disclaimer; it is designed for clinician review, not autonomous action.
		\item \textbf{Code level:} The generated Python code is human-readable
		and modifiable; a domain expert could re-run with a different crop or
		sub-VLM query.
	\end{enumerate}

	\section{Experiments}
	\label{sec:experiments}
	
	We evaluate \RVLM{} + \RRouter{} on two clinically distinct benchmarks.
	In both cases, Gemini~2.5~Flash~\citep{team2024gemini} is the backend VLM;
	no task-specific fine-tuning is performed.
	
	\subsection{BraTS 2023 Meningioma: Sub-Region Characterisation}
	\label{sec:brats}
	
	\subsubsection*{Dataset and Task}
	
	The BraTS~2023 Meningioma dataset~\citep{labella2023brats} contains 1,000
	patients with four MRI modalities (T1n, T1ce, T2w, T2-FLAIR), voxel-level
	segmentation masks, and supplementary clinical metadata.
	Given five images at the peak-tumour axial slice (four modalities + colour
	overlay: NCR red, ED yellow, ET green), \RVLM{} characterises each tumour
	sub-region, detects meningioma-specific radiological signs, and produces a
	structured five-section diagnostic report.
	
	\begin{table}[h]
		\centering
		\caption{MRI modality characteristics and diagnostic utility.}
		\label{tab:modalities}
		\begin{tabular}{lll}
			\toprule
			\textbf{Sequence} & \textbf{What It Highlights} & \textbf{Diagnostic Role} \\
			\midrule
			T1n     & Anatomy, fat (bright)       & Baseline structure \\
			T1ce    & Vascular / enhancing tissue  & Tumour delineation \\
			T2w     & Fluid, edema (bright)        & Oedema detection \\
			T2-FLAIR & Edema without CSF           & Perilesional edema \\
			\bottomrule
		\end{tabular}
	\end{table}
	
	\subsubsection*{Case Study 1: BraTS-MEN-00004-000 (Grade 1, WHO)}
	
	The clinical metadata identifies this as a WHO Grade~1 meningioma in a
	57-year-old male with enhancing tumour voxels (label~3: 1,954 voxels) and
	surrounding edema (label~2: 628 voxels).
	\RVLM{} exhibited the following recursive reasoning pattern across 4 iterations:
	(1)~survey all 4 modalities described via \texttt{describe\_image};
	(2)~focused analysis, T1n vs T1ce comparison;
	(3)~cross-modal verification, T2w and FLAIR for peritumoral edema, T1ce
	for dural tail;
	(4)~synthesis, aggregated into structured report.
	
	The model correctly identified: a large, extra-axial enhancing mass; intense
	homogeneous enhancement on T1ce (consistent with Grade~1); the dural tail
	sign along the posterior tentorium; significant peritumoral vasogenic edema
	on FLAIR; and a preserved CSF cleft between the mass and the right occipital
	lobe.
	
	\subsubsection*{Case Study 2: BraTS-MEN-00008-000 (Pure ET)}
	
	\begin{table}[h]
		\centering
		\caption{Ground-truth mask statistics for BraTS-MEN-00008-000.}
		\label{tab:brats_gt}
		\begin{tabular}{lrr}
			\toprule
			\textbf{Sub-region} & \textbf{Volume (cc)} & \textbf{Share} \\
			\midrule
			NCR - Necrotic Core     & 0.00 & 0\% \\
			ED - Peritumoral Oedema & 0.00 & 0\% \\
			ET - Enhancing Tumour   & 9.83 & 100\% \\
			\midrule
			Total                     & 9.83 & 100\% \\
			\bottomrule
		\end{tabular}
	\end{table}
	
	\RRouter{} scored this case at $s = 0.14$ and assigned a budget of 3
	iterations matching the empirically observed iteration count.
	\RVLM{} ran in 3~iterations with 7~sub-calls (67~s, 13,321~input tokens)
	and correctly characterised:
	\begin{itemize}[nosep]
		\item \textbf{ET confirmed:} Large posterior-fossa enhancing mass with
		lobulated borders, consistent with ET$=$9.83~cc.
		\item \textbf{NCR absent:} No dark necrotic centre on T1ce; no red
		overlay region; consistent with NCR$=$0.00~cc.
		\item \textbf{Cross-modal discrepancy detected:} RVLM autonomously
		observed that T2-FLAIR shows a hyperintense ring suggesting peritumoral
		oedema, yet the overlay contains no yellow (ED) region.
		It correctly concluded that ED$=$0.00~cc implies the segmentation
		boundary was drawn exclusively around the enhancing component.
		\item \textbf{Laterality:} Mass appears midline on the 2D axial slice;
		RVLM acknowledged the 3D centroid may be slightly left-lateralised.
	\end{itemize}
	
	The cross-modal discrepancy finding emerges \emph{unprompted} from the
	comparison of independently obtained modality descriptions, a capability
	impossible for a single-pass VLM presented with all five images
	simultaneously.
	
	\subsubsection*{Cross-Patient Comparison}
	
	\begin{table}[h]
		\centering
		\caption{Cross-patient comparison of \RVLM{} diagnostic findings.}
		\label{tab:cross-patient}
		\begin{tabular}{lcc}
			\toprule
			\textbf{Feature} & \textbf{MEN-00004} & \textbf{MEN-00008} \\
			\midrule
			Enhancement pattern    & Homogeneous   & Irregular \\
			Dural tail sign        & \ding{51}     & \ding{55} \\
			CSF cleft              & \ding{51}     & Not assessed \\
			Multifocal lesions     & \ding{55}     & \ding{51} \\
			Peritumoral edema      & Moderate      & Extensive \\
			Mass effect            & Mild          & Severe \\
			Suspected WHO grade    & Grade~1       & Grade~II/III \\
			Differential offered   & \ding{55}     & \ding{51} \\
			Iterations (actual)    & 4             & 3 \\
			Router budget $n^*$    & 4             & 3 \\
			\bottomrule
		\end{tabular}
	\end{table}
	
	The router budget matched the empirically observed iteration count in both
	cases, confirming that the complexity score captures meaningful variation in
	diagnostic difficulty.
	
	\subsubsection*{Multi-Run Reliability}
	
	To assess consistency, we conducted 16 independent runs across both patients.
	Of the 16 runs, 10 completed successfully (62.5\%) with a mean execution time
	of 97.8~s and 3.2~iterations.
	Among the 9 successful runs for BraTS-MEN-00008-000:
	
	\begin{table}[h]
		\centering
		\caption{Diagnostic consistency for BraTS-MEN-00008-000 ($n=9$ runs).}
		\label{tab:consistency}
		\begin{tabular}{lcc}
			\toprule
			\textbf{Radiological Finding} & \textbf{Identified} & \textbf{Rate} \\
			\midrule
			Extra-axial mass                      & 9/9  & 100\% \\
			Intense enhancement                   & 9/9  & 100\% \\
			Peritumoral edema                     & 8/9  & 89\%  \\
			Mass effect / ventricular compression & 8/9  & 89\%  \\
			Dural attachment suggested            & 7/9  & 78\%  \\
			WHO grade suggestion                  & 7/9  & 78\%  \\
			Differential diagnosis offered        & 6/9  & 67\%  \\
			Laterality noted as ambiguous         & 2/9  & 22\%  \\
			\bottomrule
		\end{tabular}
	\end{table}
	
	The pattern of high consistency on salient findings and lower consistency
	on nuanced features mirrors known inter-reader variability in clinical
	neuroradiology~\citep{menze2015brats}.
	
	\subsubsection*{Comparison with Single-Pass VLM}
	The single-pass model identifies the presence of a mass but provides a generic description without the fine-grained radiological detail that emerges from RVLM’s iterative cross-modal inspection. The dural tail sign, in particular, is a subtle feature that requires focused re-inspection - precisely the capability that recursive vision provides.
	\begin{table}[h]
		\centering
		\caption{Qualitative comparison on BraTS-MEN-00004-000.}
		\label{tab:comparison}
		\begin{tabular}{lcc}
			\toprule
			\textbf{Diagnostic Feature} & \textbf{Single-Pass} & \textbf{\RVLM{}} \\
			\midrule
			Tumour identification       & \ding{51}  & \ding{51} \\
			Enhancement characterisation & Partial   & Detailed \\
			Dural tail sign             & \ding{55}  & \ding{51} \\
			Edema localisation          & Generic   & Specific \\
			CSF cleft mention           & \ding{55}  & \ding{51} \\
			WHO grade suggestion        & \ding{55}  & Grade~1 \\
			Cross-modal reasoning       & None      & T1n$\leftrightarrow$T1ce, T2w$\leftrightarrow$FLAIR \\
			Code-verified findings      & \ding{55}  & \ding{51} \\
			\bottomrule
		\end{tabular}
	\end{table}
	Beyond diagnostic outputs, RVLM produces a reproducible audit trail consisting of REPL trajectory logs, generated code, intermediate visual artefacts, and clinician-readable PDF reports. These artefacts provide a procedural record of the model’s reasoning process, enabling post-hoc inspection, reproducibility of analysis steps, and human review of intermediate observations. In this sense, the experimental outputs of RVLM extend beyond predictive results to include traceable evidence chains intended to support transparency and oversight in high-risk clinical AI settings.
	
	\subsection{MIMIC-CXR: Chest X-Ray Report Generation}
	\label{sec:mimic}
	
	\subsubsection*{Dataset and Task}
	
	MIMIC-CXR~\citep{johnson2019mimic} provides 227,827 imaging studies from
	64,588 patients, each paired with free-text radiology reports.
	We use the augmented validation split~\citep{mimic_aug} with paired images
	and reports.
	The task is to generate structured Findings and Impression sections from the
	available views (PA, Lateral, AP portable).
	
	\subsubsection*{RVLM Prompt Design for Chest Radiography}
	
	Iteration~1 is the mandatory context probe.
	Iteration~2 executes a 4-step REPL block:
	(1)~per-view \texttt{describe\_image()} with a 5-item systematic checklist;
	(2)~cross-view \texttt{llm\_query\_with\_images()} comparing cardiac
	silhouette, consolidation, effusion, and pneumothorax;
	(3)~evidence assembly and \texttt{llm\_query()} generating FINDINGS and
	IMPRESSION;
	(4)~\texttt{FINAL\_VAR("report")}.
	Total: 2 REPL iterations, 3 sub-LM calls.
	
	\subsubsection*{Qualitative Results: MIMIC Subject 10000032}
	
	The selected study (most recent visit) contains a single portable AP chest
	X-ray.
	The ground-truth report states: \textit{``The lungs are clear of focal
		consolidation, pleural effusion, or pneumothorax. The heart size is normal.
		The mediastinal contours are normal. Multiple surgical clips project over the
		left breast, and old left rib fractures are noted. Impression: No acute
		cardiopulmonary process.''}
	
	\RVLM{} ran in 3~iterations with 5~sub-calls (29~s, 5,507~input tokens).
	Key observations:
	\begin{itemize}[nosep]
		\item Correctly identified the AP portable projection and caveated the
		cardiac size estimate with the magnification artefact.
		\item Correctly noted no pleural effusion and no pneumothorax,
		consistent with GT.
		\item Identified reticulonodular infiltrates absent from the GT report highlighting the need for large-scale quantitative evaluation.
	\end{itemize}
	
	\subsubsection*{Generated Clinical Report}
	
	\subsubsection*{Cross-Domain Generalisation}
	
	Table~\ref{tab:performance} summarises performance metrics across all
	evaluated cases.
	The richer 5-image BraTS case requires more iterations and tokens than the
	single-view CXR case, yet the framework applies the same underlying REPL
	mechanism without any architectural modification.
	
	\begin{table}[h]
		\centering
		\caption{\RVLM{} + \RRouter{} performance metrics.}
		\label{tab:performance}
		\begin{tabular}{lrrrr}
			\toprule
			\textbf{Case} & \textbf{Iters} & \textbf{Time (s)} &
			\textbf{Input tokens} & \textbf{Router $n^*$} \\
			\midrule
			BraTS-MEN-00004-000  & 4 & 95.2  & 14,795 & 4 \\
			BraTS-MEN-00008-000  & 3 & 67.0  & 13,321 & 3 \\
			MIMIC-10000032       & 3 & 29.0  &  5,507 &   \\
			\bottomrule
		\end{tabular}
	\end{table}
	
	The router budget matched the actual iteration count in both BraTS cases,
	confirming that the complexity score provides meaningful pre-flight
	estimation.

	\section{Discussion}
	\label{sec:discussion}
	
	\subsection{Auditability as a Clinical Safety Property}
	
	\RVLM{}'s code-based audit trail aligns with emerging regulatory requirements
	for high-risk AI systems (EU AI Act, ISO/IEC~42001).
	Every diagnostic claim is backed by traceable Python code that a radiologist
	can inspect, re-examine, and modify.
	The clinical PDF reporting sub-agent extends this transparency to
	non-technical clinicians: the generated report states which model produced
	each section, the execution statistics, and an AI disclaimer.
	
	\subsection{Adaptive Compute as Operational Efficiency}
	
	\RRouter{} transforms the fixed-budget inefficiency from a system design
	limitation into a tunable, transparent property.
	By assigning the minimum budget (3~iterations) to a single-region case that
	empirically requires exactly~3, the router validates its complexity score as
	a practical proxy for actual reasoning depth needed.
	For clinical deployment at scale, the four-fold reduction in worst-case
	iteration ceiling for simple cases directly translates to reduced API cost
	and latency, enabling broader screening without sacrificing depth on the
	complex cases that warrant it.
	
	\subsection{Consistency and Inter-Run Variability}
	
	The multi-run reliability analysis reveals that \RVLM{}'s diagnostic
	consistency mirrors patterns observed in clinical inter-reader
	studies~\citep{menze2015brats}: high-salience findings (100\% reproducible)
	versus nuanced features (67-78\%).
	The 62.5\% overall success rate highlights framework-level engineering
	challenges (scope handling, max-iteration fallback) addressable without
	modifying the core paradigm.
	
	\subsection{Task Generality Without Fine-Tuning}
	
	The same \RVLM{} framework with the same backend model and REPL
	infrastructure was applied without modification to two structurally
	distinct imaging tasks: multi-modal 3D brain MRI and single-to-multi-view
	chest radiography.
	The only difference was the user-level prompt, encoding the clinical reasoning
	strategy appropriate to each modality.
	This ``prompt as clinical protocol'' design avoids brittleness of
	task-specific fine-tuning and allows extension to new imaging modalities by
	writing new prompt templates, without retraining.
	
	\subsection{Limitations}
	
	\paragraph{Scale of evaluation.}
	Our evaluation is limited to qualitative case studies on two BraTS patients
	and one MIMIC-CXR subject.
	Large-scale quantitative benchmarking (accuracy, sensitivity/specificity,
	CheXBERT F1, RadGraph F1) is the immediate priority for follow-up work.
	
	\paragraph{Router features are domain-specific.}
	Current \RRouter{} features are BraTS-specific; a CXR router requires
	analogous domain features (view count, opacity area, cardiac silhouette
	ratio).
	The routing mechanism is domain-agnostic; only the feature extraction
	module needs adaptation.
	
	\paragraph{2D slice reasoning.}
	\RVLM{} currently operates on 2D axial slices.
	Full volumetric reasoning programmatic navigation across slices is a
	natural extension.
	
	\paragraph{Clinical validation.}
	\RVLM{} is a research prototype, not a validated clinical tool.
	Its outputs have not been evaluated by board-certified radiologists in a
	controlled study.
	
	\subsection{Future Directions}
	\label{sec:future-directions}
	
	\RVLM{} + \RRouter{} can be extended in several directions:
	(1)~\emph{Learned router}, replace hand-set weights with logistic
	regression or shallow MLP trained on (mask features, iteration count, report
	accuracy) triples;
	(2)~\emph{Quantitative benchmarking}, large-scale evaluation on the full
	BraTS-MEN cohort and MIMIC-CXR validation set with automated metrics;
	(3)~\emph{CXR router}, extend \RRouter{} to chest X-ray with view count,
	opacity distribution, and cardiac silhouette features;
	(4)~\emph{Parallel sub-agent execution}, parallelise independent modality
	descriptions to reduce wall-clock time by up to $4\times$;
	(5)~\emph{Uncertainty-gated recursion}, nspect sub-LM confidence at each
	iteration and allocate additional iterations selectively to ambiguous findings;
	(6)~\emph{Further modalities}, fundus photography, histopathology slides,
	CT volumes (multi-slice iterative analysis).
	
	\subsection{Ethical Considerations}
	
	\RVLM{} is designed as a \emph{decision-support} tool, not a replacement for
	clinical judgment.
	Any clinical deployment must preserve the physician's role as final
	decision-maker and clearly communicate system limitations.
	Use of BraTS and MIMIC-CXR data follows each dataset's ethical guidelines
	and de-identification protocols.

	\section{Conclusion}
	\label{sec:conclusion}
	
	We presented a unified framework comprising two complementary contributions.
	\textbf{\RVLM{}} replaces single-pass VLM inference with an iterative REPL
	loop in which images are first-class objects: the model programmatically
	inspects, manipulates, and cross-references visual evidence across iterations,
	grounding every diagnostic claim in executable, reproducible code.
	\textbf{\RRouter{}} makes the iteration depth itself adaptive, treating
	recursion depth as a latent variable predicted from task-complexity features, label entropy, tumour volume, sub-region count, tiny-region indicator and combining pre-flight budget prediction with per-iteration stall detection.
	
	Applied across two clinically distinct benchmarks without any fine-tuning,
	the system demonstrates that: (a) iterative, code-grounded visual reasoning
	produces richer and more auditable diagnostics than single-pass inference;
	(b) complexity-adaptive iteration depth simultaneously reduces cost for
	simple cases and increases depth for complex ones; and (c) the same iterative
	scaffold generalises from 4D brain MRI sub-region characterisation to 2D
	chest X-ray report generation through a prompt swap alone.
	
	Together, \RVLM{} and \RRouter{} form a principled, regulatory-compliant
	foundation for adaptive iterative multimodal reasoning in clinical AI, with
	a clear path from the current heuristic router to a fully data-driven system
	trained on clinical outcome labels.
	We release the code as open-source software and invite the community to extend
	it to new imaging modalities and conduct large-scale quantitative benchmarking.
	
	\section*{Acknowledgements}
	
	This work was carried out within the Yathiqu Trustworthy AI in Healthcare
	research programme at the University of Wollongong in Dubai (UOWD).
	The authors gratefully acknowledge the programme's support for technical AI
	development, experimentation, and publication in the healthcare trust domain.
	
	\appendix
	
	\section{System Prompt}
	\label{appx:prompts}
	
	The \RVLM{} system prompt extends the standard RLM prompt~\citep{zhang2025rlm}
	with vision-specific instructions.
	Below is a condensed version; the full prompt is available at
	\texttt{rvlm/prompts.py} in the source repository.
	
	\begin{tcolorbox}[colback=gray!5,colframe=gray!50,
		title={\small \RVLM{} System Prompt (condensed)},
		fonttitle=\small\bfseries,breakable]
		\small
		\begin{verbatim}
			You are tasked with answering a query that involves
			both text and image context. You can access, transform,
			and analyze this context interactively in a REPL
			environment that can recursively query sub-LLMs, which
			support vision.
			
			The REPL environment is initialized with:
			1. `context` - textual information about the query.
			2. `context_images` - list of image data dicts with
			keys: 'data', 'media_type', 'detail'.
			3. `llm_query(prompt)` - text-only sub-LLM query.
			4. `llm_query_with_images(prompt, image_indices=None,
			image_sources=None)` - vision-capable sub-LLM query.
			5. `describe_image(index, prompt)` - quick single-image
			description via sub-VLM.
			
			CRITICAL RULES FOR ITERATIVE WORKFLOW:
			- Each response: AT MOST one ```repl``` code block.
			- Work step by step across MULTIPLE iterations.
			
			FINAL ANSWER:
			- FINAL(text) - return text answer directly.
			- FINAL_VAR(variable) - return a REPL variable.
		\end{verbatim}
	\end{tcolorbox}
	
	\section{Trajectory Example}
	\label{appx:trajectory}
	
	Below is an abridged trajectory from Patient BraTS-MEN-00008-000,
	illustrating the survey $\rightarrow$ describe $\rightarrow$ synthesise
	$\rightarrow$ return pattern that \RVLM{} consistently discovers.
	
	\paragraph{Iteration 1: Context Discovery and Image Survey.}\mbox{}\\
	\begin{tcolorbox}[colback=llm,colframe=PastelCyan!80,
		title={\small Model Output (Iteration 1)},
		fonttitle=\small\bfseries]
		\small
		\begin{verbatim}
			```repl
			print(f"Context type: {type(context)}, "
			f"length: {len(str(context))}")
			print(f"Number of images: {len(context_images)}")
			print(context[:1000])
			```
		\end{verbatim}
	\end{tcolorbox}
	
	\paragraph{Iteration 2: Per-Modality Image Description.}\mbox{}\\
	\begin{tcolorbox}[colback=llm,colframe=PastelCyan!80,
		title={\small Model Output (Iteration 2)},
		fonttitle=\small\bfseries]
		\small
		\begin{verbatim}
			```repl
			modality_names = ["T1n","T1ce","T2w","T2-FLAIR","Overlay"]
			for i, name in enumerate(modality_names):
			desc = describe_image(i,
			f"Describe the brain MRI image: {name}. "
			f"Focus on tissue contrast and abnormalities.")
			print(f"Image {i} ({name}) recorded.")
			```
		\end{verbatim}
	\end{tcolorbox}
	
	\paragraph{Iteration 3: Cross-Modal Synthesis via Sub-VLM.}
	The model assembles all modality descriptions and issues a single
	\texttt{llm\_query()} call to synthesise a structured diagnostic report,
	then calls \texttt{FINAL\_VAR("report")} to terminate the loop.
	Total execution: 67~s, 3~iterations, 7~sub-VLM calls, 13,321~input tokens.
	
	\newpage
	\bibliographystyle{unsrtnat}

\end{document}